\documentclass[final,5p,times,twocolumn,nopreprintline]{elsarticle}

\usepackage{amsmath,amssymb,amsfonts}
\usepackage{algorithmic}
\usepackage{graphicx,color}
\usepackage{textcomp}
\usepackage{hyperref}
\usepackage{balance}

\usepackage{graphicx}
\usepackage{tabularx}
\usepackage{makecell}
\usepackage[table]{xcolor}
\usepackage{booktabs}
\usepackage{listings}
\usepackage{xcolor}
\usepackage{array}
\usepackage{xstring}
\usepackage{subfig}
\usepackage{stfloats}
\usepackage{tikz} \usetikzlibrary{calc, arrows.meta, intersections, patterns, positioning, shapes.misc, fadings, through,decorations.pathreplacing,shadows.blur, decorations.pathmorphing}
\colorlet{punct}{red!60!black}
\definecolor{background}{HTML}{EEEEEE}
\definecolor{delim}{RGB}{20,105,176}
\colorlet{numb}{magenta!60!black}
\usepackage{etoolbox}

\makeatletter
\renewcommand{\thesubsubsection}{\arabic{subsubsection})}

\renewcommand\@seccntformat[1]{
  \ifstrequal{#1}{subsubsection}
    {\thesubsubsection\quad}
    {\csname the#1\endcsname.\quad}
}
\makeatother

\def\BibTeX{{\rm B\kern-.05em{\sc i\kern-.025em b}\kern-.08em
    T\kern-.1667em\lower.7ex\hbox{E}\kern-.125emX}}
\AtBeginDocument{\definecolor{ojcolor}{cmyk}{0.93,0.59,0.15,0.02}}

\usepackage{fancyhdr}
\begin{document}

\begin{frontmatter}

\title{Video Dataset for Surgical Phase, Keypoint, and Instrument\\ Recognition in Laparoscopic Surgery (PhaKIR)}

\author[1,2]{Tobias Rueckert\corref{cor1}}
    \cortext[cor1]{
    Corresponding authors.\\
    \indent \indent \hspace{-1.2mm} \textit{E-mail addresses:}\\ \href{mailto:tobias.rueckert@oth-regensburg.de}{tobias.rueckert@oth-regensburg.de}; \href{mailto:christoph.palm@oth-regensburg.de}{christoph.palm@oth-regensburg.de} \\
    This work has been submitted to the IEEE for possible publication. Copyright may be transferred without notice, after which this version may no longer be accessible.
        }
    \author[1]{Raphaela Maerkl}
    \author[1]{David Rauber}
    \author[1,3,4]{Leonard Klausmann}
    \author[1]{Max Gutbrod}
    \author[5,6]{Daniel Rueckert}
    \author[7,8]{Hubertus Feussner}
    \author[7,8]{Dirk Wilhelm}
    \author[1,3,4]{Christoph Palm\corref{cor1}}
    
    \address[1]{Regensburg Medical Image Computing (ReMIC), OTH Regensburg, Regensburg, Germany}
    \address[2]{AKTORmed Robotic Surgery, Neutraubling, Germany}
    \address[3]{Regensburg Center of Biomedical Engineering (RCBE), OTH Regensburg and Regensburg University, Regensburg, Germany}
    \address[4]{Regensburg Center of Health Sciences and Technology (RCHST), OTH Regensburg, Regensburg, Germany}
    \address[5]{Chair for AI in Healthcare and Medicine, Technical University of Munich (TUM) and TUM University Hospital, Munich, Germany}
    \address[6]{Biomedical Image Analysis Group, Department of Computing, Imperial College London, London, UK}
    \address[7]{Research Group MITI, TUM University Hospital, School of Medicine and Health, Technical University of Munich, Munich, Germany}
    \address[8]{Department of Surgery, TUM University Hospital, School of Medicine and Health, Technical University of Munich, Munich, Germany}

\begin{abstract}
Robotic- and computer-assisted minimally invasive surgery (RAMIS) is increasingly relying on computer vision methods for reliable instrument recognition and surgical workflow understanding.  
Developing such systems often requires large, well-annotated datasets, but existing resources often address isolated tasks, neglect temporal dependencies, or lack multi-center variability.  

We present the \textit{Surgical Procedure Phase, Keypoint, and Instrument Recognition (PhaKIR)} dataset, comprising eight complete laparoscopic cholecystectomy videos recorded at three medical centers.  
The dataset provides frame-level annotations for three interconnected tasks: surgical phase recognition (485{,}875 frames), instrument keypoint estimation (19{,}435 frames), and instrument instance segmentation (19{,}435 frames).  
PhaKIR is, to our knowledge, the first multi-institutional dataset to jointly provide phase labels, instrument pose information, and pixel-accurate instrument segmentations, while also enabling the exploitation of temporal context since full surgical procedure sequences are available.
It served as the basis for the PhaKIR Challenge as part of the Endoscopic Vision (EndoVis) Challenge at MICCAI 2024 to benchmark methods in surgical scene understanding, thereby further validating the dataset’s quality and relevance.
The dataset is publicly available upon request via the Zenodo platform.
\end{abstract}

 
\end{frontmatter}
\section*{BACKGROUND}

\begin{figure*}[t!]
    \centering
    \includegraphics[width=0.8\linewidth]{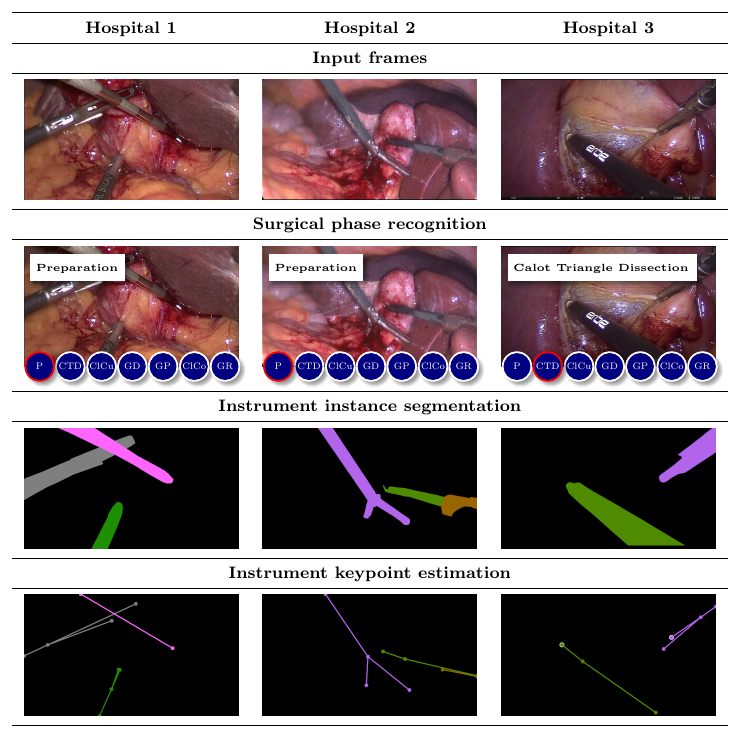}
    \caption{Overview of the PhaKIR dataset, illustrating source video data and the annotations for the three tasks: surgical phase recognition, instrument instance segmentation, and instrument keypoint estimation, across three medical centers.}
    \label{fig:annotations}
\end{figure*}

Minimally invasive surgery (MIS) offers advantages over open procedures, such as reduced invasiveness, faster recovery, shorter hospital stays, and a lower risk of postoperative infection, making it the standard approach for many interventions, including cholecystectomy~\cite{darzi2002recent}, \cite{de2018minimally}, \cite{van2012robot}.  
Robotic- and computer-assisted minimally invasive surgery (RAMIS) aims to further support the surgical team, often through machine learning-based methods~\cite{haidegger2022robot}, \cite{maier2022surgical}.  
The development of such systems critically depends on reliable recognition and spatial localization of surgical instruments, combined with contextual interpretation of the surgical workflow~\cite{rueckert2024methods}, \cite{twinanda2016endonet}.  
Instrument recognition can be achieved by segmentation of the instruments or by estimating keypoints to approximate tool poses, while workflow context can be inferred through surgical phase recognition.  

\begin{table*}[t]
    \small
    \centering
    \caption{Video numbers together with their source hospital, duration in minutes, number of raw frames, and number of annotated frames for the phase recognition task, as well as for the instrument instance segmentation and keypoint estimation tasks. Videos are not numbered consecutively because intermediate recordings were extracted for the test dataset. }
    \label{tab:videos}
    \begin{tabularx}{\textwidth}{l *{8}{>{\centering\arraybackslash}X} >{\bfseries}c}
        \toprule
        Video & 1 & 2 & 3 & 4 & 5 & 7 & 11 & 13 & Total \\
        \midrule
        Hospital & MRI & MRI & MRI & MRI & MRI & MRI & UKHD & KWS &  \\
        Duration (min) & 39:07 & 48:12 & 32:39 & 49:10 & 28:31 & 57:03 & 32:54 & 36:19 & 323:55 \\
        \#Frames (raw) & 58,675 & 72,300 & 48,975 & 73,750 & 42,775 & 85,575 & 49,350 & 54,475 & 486,875 \\
        \#Annotated (phase) & 58,675 & 72,300 & 48,975 & 73,750 & 42,775 & 85,575 & 49,350 & 54,475 & 486,875 \\
        \#Annotated (seg., kp.) & 2,347 & 2,892 & 1,959 & 2,950 & 1,711 & 3,423 & 1,974 & 2,179 & 19,475 \\
        \bottomrule
    \end{tabularx}
\end{table*}

High-quality, publicly available datasets are crucial for training and benchmarking these assistance systems.  
Although several datasets exist, they usually address only isolated tasks such as surgical phase recognition~\cite{twinanda2016endonet}, \cite{stauder2016tum}, \cite{wagner2023comparative}, instrument segmentation~\cite{bodenstedt2015comparative}, \cite{allan2017robotic}, \cite{allan2018robotic}, \cite{ross2019comparative}, \cite{zia2021objective}, \cite{psychogyios2022sar}, \cite{nwoye2022cholec}, \cite{zia2022surgical}, \cite{malpani2023synthetic}, \cite{bodenstedt2023}, or keypoint estimation~\cite{bodenstedt2015comparative}.  
Additionally, they only partially reflect the real-world requirements of surgical procedures.  
For example, only one dataset includes data from multiple medical centers~\cite{wagner2023comparative}.  
Moreover, instrument types~\cite{bodenstedt2015comparative}, \cite{allan2018robotic}, \cite{ross2019comparative}, \cite{malpani2023synthetic} and individual instances~\cite{bodenstedt2015comparative}, \cite{allan2017robotic}, \cite{allan2018robotic}, \cite{psychogyios2022sar}, \cite{malpani2023synthetic} are often not distinguished, complete procedures are rarely provided~\cite{allan2017robotic}, \cite{allan2018robotic}, \cite{ross2019comparative}, \cite{bodenstedt2023}, and some datasets are based on non-human tissue~\cite{allan2017robotic}, \cite{allan2018robotic}, \cite{malpani2023synthetic}. 
These limitations hinder temporal modeling, generalization, and clinical realism.

To address these gaps, we present the \textit{Surgical Procedure \textbf{Pha}se, \textbf{K}eypoint, and \textbf{I}nstrument \textbf{R}ecognition (PhaKIR)} dataset, which served as the training resource for the PhaKIR Challenge~\cite{rueckert2025comparative}, a sub-challenge of the Endoscopic Vision Challenge (EndoVis) at MICCAI 2024.  
The dataset comprises eight complete laparoscopic cholecystectomy videos collected from three medical centers, capturing real surgeries and thus enabling both temporal modeling and the study of inter-institutional variability.  
We provide unified annotations for three interconnected tasks, illustrated in Figure~\ref{fig:annotations}: surgical phase recognition, instrument instance segmentation, and instrument keypoint estimation.  
This combination allows researchers to determine instrument type, location, and instance while simultaneously modeling procedural context through phase recognition. 

The dataset is publicly available upon request via the Zenodo platform~\cite{rueckert2025dataset}.

\section*{COLLECTION METHODS AND DESIGN} 

The creation of the PhaKIR dataset followed a structured workflow, including video recording (see~Sect.~\ref{methods:recording}), annotation for three tasks (see~Sec.~\ref{methods:annotation}), and preparation of the challenge dataset (see~Sec.~\ref{methods:challenge}).  

\subsection{Recording of surgical videos}
\label{methods:recording}

\begin{figure*}[b!]
    \centering
    \includegraphics[width=1.0\linewidth]{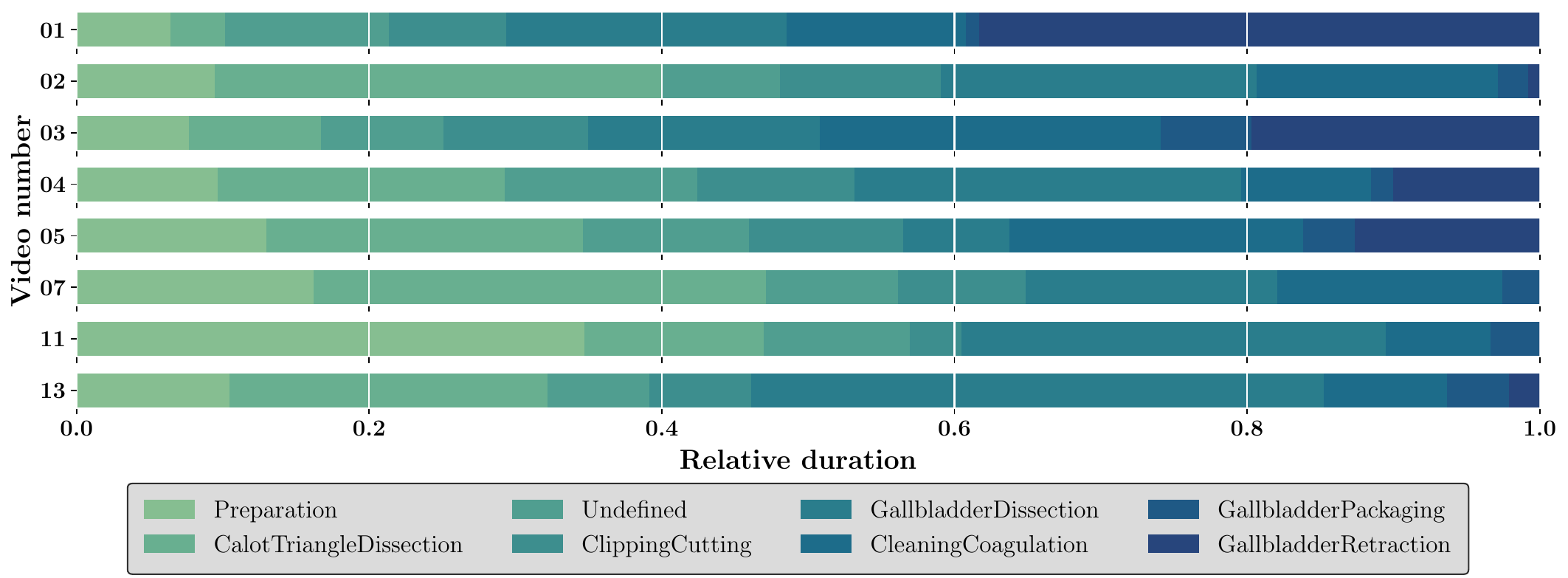}
    \caption{Visualization of the relative duration of each surgical phase for each video. The phases are arranged in order of their most frequent occurrence across all recorded interventions.}
    \label{fig:phase_durations}
\end{figure*}

A total of $n=8$ videos, with durations ranging from 28 to 58 minutes, were recorded during routine minimally invasive cholecystectomies on adult patients of different genders.  
Recordings were acquired with different endoscopic cameras at 25 frames per second (fps) and a resolution of $1920 \times 1080$ pixels.  
The videos originated from three German hospitals: TUM University Hospital Rechts der Isar~(MRI,~\cite{MRI_TUM}) ($n=6$), Heidelberg University Hospital~(UKHD,~\cite{UKHD}) ($n=1$), and Weilheim-Schongau Hospital~(KWS,~\cite{MeinKrankenhaus2030}) ($n=1$). 
The Heidelberg video corresponds to \texttt{HeiChole2.mp4}, previously published as part of the Surgical Workflow and Skill Analysis Challenge (HeiChole Benchmark, ~\cite{wagner2023comparative}), and was re-annotated for consistency.
Sequences showing regions outside the abdominal cavity were removed to protect patient and staff privacy; cut indices are provided for each video in \texttt{Video\_xx\_Cuts.csv}.  
An overview of video sources, durations, and annotation counts is given in Table~\ref{tab:videos}.

\subsection{Annotation process}
\label{methods:annotation}

\begin{figure*}[!t]
\centering

\subfloat[][]{\includegraphics[scale=0.25]{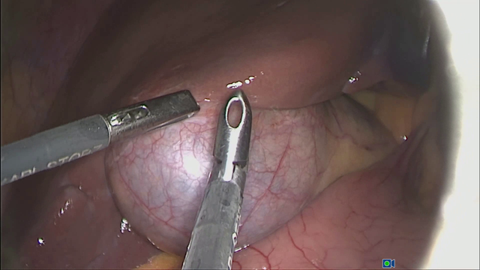}\label{instruments_000}}
\hfill
\subfloat[][]{\includegraphics[scale=0.25]{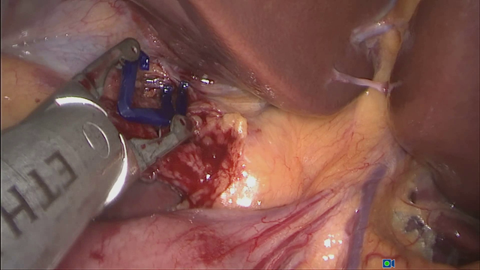}\label{instruments_001}}
\hfill
\subfloat[][]{\includegraphics[scale=0.25]{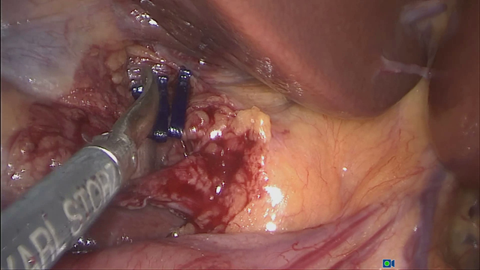}\label{instruments_002}}
\hfill
\subfloat[][]{\includegraphics[scale=0.25]{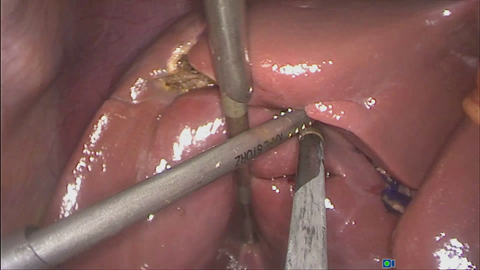}\label{instruments_003}}\\ 
\subfloat[][]{\includegraphics[scale=0.25]{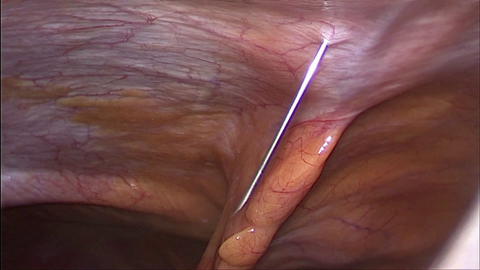}\label{instruments_004}}
\hfill
\subfloat[][]{\includegraphics[scale=0.25]{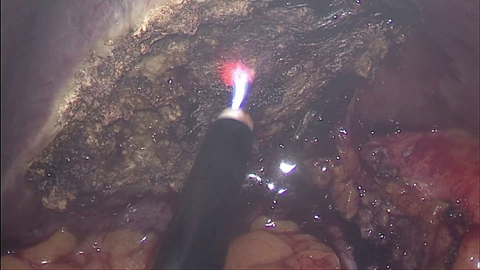}\label{instruments_005}}
\hfill
\subfloat[][]{\includegraphics[scale=0.25]{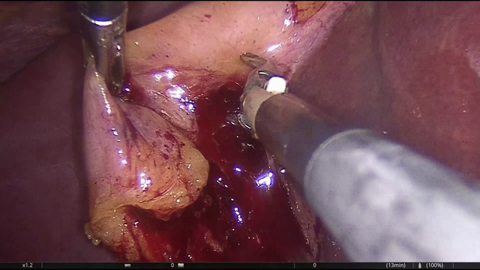}\label{instruments_006}}
\hfill
\subfloat[][]{\includegraphics[scale=0.25]{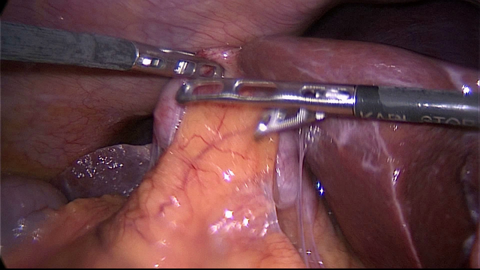}\label{instruments_007}}\\ 
\hfill
\subfloat[][]{\includegraphics[scale=0.25]{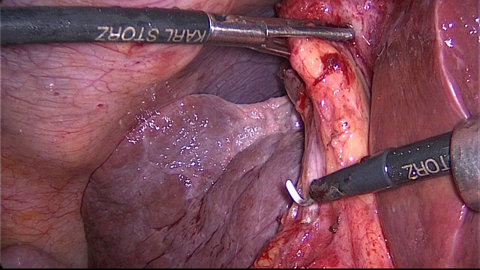}\label{instruments_008}}
\hfill
\subfloat[][]{\includegraphics[scale=0.25]{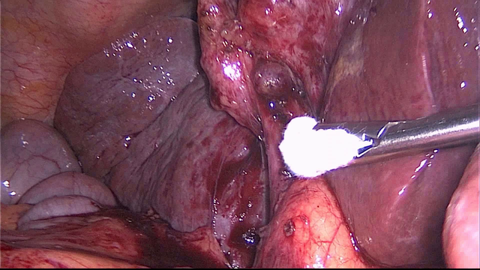}\label{instruments_009}}
\hfill 
\subfloat[][]{\includegraphics[scale=0.25]{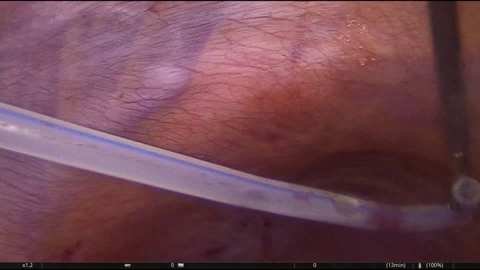}\label{instruments_010}}
\hfill
\subfloat[][]{\includegraphics[scale=0.25]
{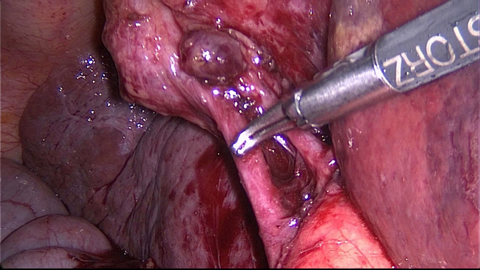}\label{instruments_011}}\\

{\caption{Instrument types included in the dataset. In case of more than one instrument in an image, the instruments are listed from left to right in the order they appear: Grasper, PE-Forceps (a), Clip-Applicator (b), Scissor (c), Trocar-Tip, Suction-Rod, Palpation-Probe, HF-Coag.-Probe (d), Needle-Probe (e), Argonbeamer (f), Blunt-Grasper-Spec., Bipolar-Clamp (g), Blunt-Grasper, Blunt-Grasper-Curved (h), Blunt-Grasper-Spec., Dissection-Hook, Trocar-Tip (i), Sponge-Clamp (j), Drainage (k), Overholt (l).} \label{fig:instruments}}

\end{figure*}

Annotations were created by four annotators with medical expertise (one senior surgeon and three medically trained students).  
Following the description of the frame extraction process, each task is described in terms of its objective, manual annotation process, and ground-truth generation.  

\subsubsection{Frame extraction}
\label{methods:annotation:frameextraction}

For surgical phase recognition, every frame was annotated (25 fps), resulting in 485,875 labeled frames.  
For instrument instance segmentation and instrument keypoint estimation, one frame per second was annotated (every 25th frame), resulting in 19,435 labeled images per task.  
All tasks cover the identical and complete surgical sequences.  

\subsubsection{Task 1: Surgical phase recognition}
\label{methods:annotation:phase}

\textbf{Objective} Each frame was assigned to one of seven phases defined in the Cholec80 dataset~\cite{twinanda2016endonet}, with an additional \textit{undefined} phase to capture transitions.
Examples of annotated frames are shown in the second row of Figure~\ref{fig:annotations}, while the relative phase durations across videos are visualized in Figure~\ref{fig:phase_durations}.

\noindent \textbf{Manual annotation} Timestamps marking phase transitions were documented during manual review.
The annotators had access to the full-length recordings without timeline restrictions.
The start of a phase was defined as the first appearance of characteristic instruments performing essential actions, while the phase end was defined as the disappearance of all such instruments.  
The Heidelberg video~\cite{wagner2023comparative} was fully re-annotated to ensure protocol consistency.  

\noindent \textbf{Ground-truth generation} Documented timestamps were converted into csv files.
All frames between start and end markers were automatically assigned to the corresponding phase.  
The final csv files provides frame-level phase labels for each video.

\subsubsection{Task 2: Instrument instance segmentation}
\label{methods:annotation:segmentation}

\begin{figure*}[t!]
    \centering
    \includegraphics[width=0.9\linewidth]{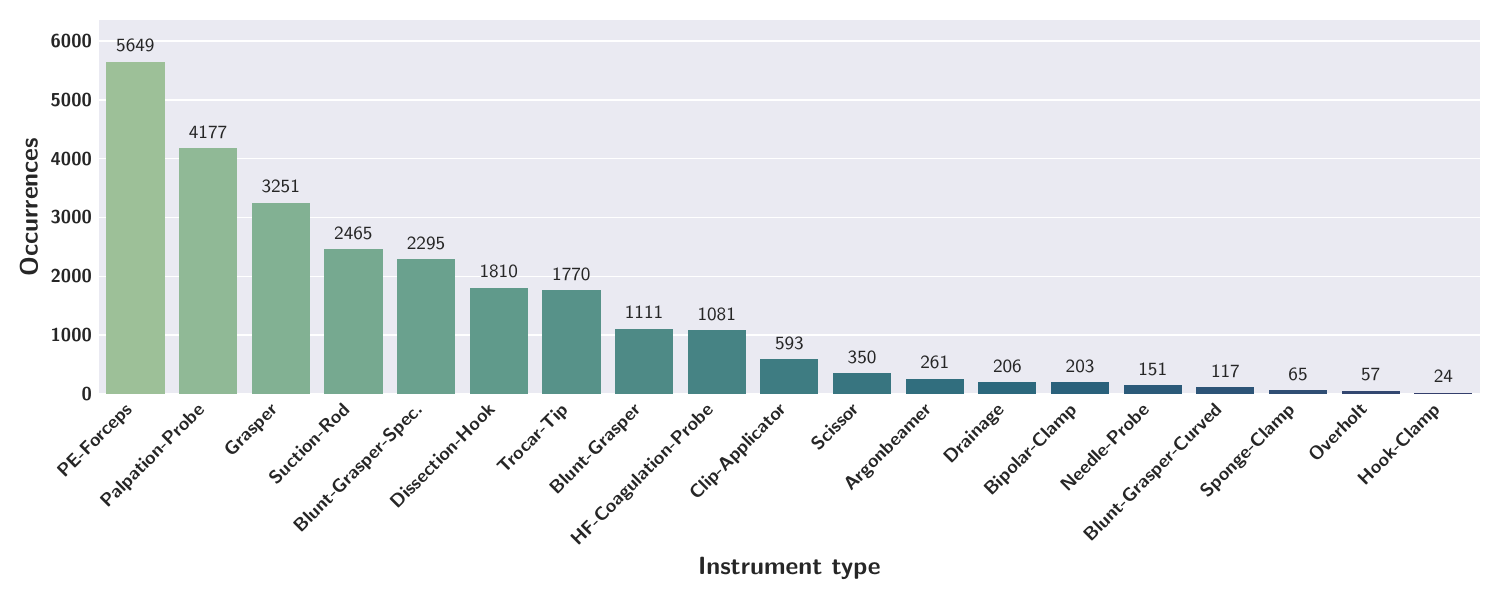}
    \caption{Number of frames in which the respective instrument type occurs.}
    \label{fig:instrument_occurences}
\end{figure*}

\textbf{Objective} Instruments were segmented in every 25th frame and assigned to one of 19 classes.
Each pixel was labeled as instrument or background.
Multiple instances of the same class were distinguished.
Examples of segmentation annotations are illustrated in the third row of Figure~\ref{fig:annotations}.
The complete set of instrument classes is shown in Figure~\ref{fig:instruments}, and their frequency of occurrence across the dataset is depicted in Figure~\ref{fig:instrument_occurences}.

\noindent \textbf{Manual annotation} Segmentation was performed using the Computer Vision Annotation Tool (CVAT, ~\cite{cvatai2023cvat}) with polygon contours.  
Annotators had access to the complete video sequences to improve accuracy, but only visible parts of instruments were labeled.  
Different instances of the same class were temporarily separated with pseudo-classes to support ground-truth generation.  

\noindent \textbf{Ground-truth generation} Manual annotations were converted into segmentation masks via Python scripts.  
Masks were encoded using three color channels: red and green channels encode the instrument class, while the blue channel encodes the instrument instance.  
Instruments of the same class share identical red and green values but differ in the blue channel (see  Table~\ref{tab:segmentation_colors}).  

\subsubsection{Task 3: Instrument keypoint estimation}
\label{methods:annotation:keypoint}

\textbf{Objective} For every 25th frame, keypoints describing instrument pose were annotated together with their visibility status (visible, occluded, or not available).  
The number of keypoints depended on the instrument type, ranging from two to four.  
Examples of keypoint annotations are presented in the last row of Figure~\ref{fig:annotations}, with hidden points marked by white circles.  

\noindent \textbf{Manual annotation} Keypoints were annotated in CVAT using the ``point'' function.  
For each instrument a set of predefined keypoints was specified: the endpoint where the instrument enters the image border, the shaft-to-tip junction marking the transition between these parts, and the instrument tip, which in the case of opening tools was represented by two separate points. 

Depending on the instrument type, between two and four keypoints were specified with the number and type of keypoints for each instrument shown in Table~\ref{tab:keypoints}.
Each keypoint was additionally labeled with a visibility status following the COCO protocol~\cite{coco_keypoints} (visible, occluded, or not available).

\begin{table}[b!]
\footnotesize
\centering
\caption{Surgical instruments grouped by the number and type of annotated keypoints. Keypoint labels are abbreviated by T1 (Tip1), T2 (Tip2), SP (ShaftPoint), and EP (EndPoint).}
\label{tab:keypoints}
\begin{tabularx}{1.0\linewidth}{@{}p{0.1\textwidth} p{0.85\textwidth}@{}}
\toprule
\textbf{Keypoints} & \textbf{Instruments} \\
\midrule
\makecell[{{l}}]{4 Keypoints \\ (T1, T2, SP, EP)} &
\makecell[{{l}}]{Bipolar-Clamp, Blunt-Grasper, Blunt-Grasper-Curved, \\
Blunt-Grasper-Spec., Clip-Applicator, Grasper, Hook-\\
Clamp, Overholt, PE-Forceps, Scissor, Sponge-Clamp} \\[0.5em]

\makecell[{{l}}]{3 Keypoints \\ (T1, SP, EP)} &
\makecell[{{l}}]{Argonbeamer, Dissection-Hook, HFcoag-Probe, \\
Suction-Rod} \\[0.5em]

\makecell[{{l}}]{2 Keypoints \\ (T1, EP)} &
\makecell[{{l}}]{Drainage, Needle-Probe, Palpation-Probe, Trocar-Tip} \\
\bottomrule
\end{tabularx}
\end{table}

Annotators were given access to the full video sequences, including both preceding and subsequent frames to support precise positioning of the keypoints.
In contrast to segmentation, however, the annotation of keypoints the temporal continuity and expected motion of the instrument were taken into account to refine the placement, particularly when the target region was partially occluded.

\noindent \textbf{Ground-truth generation} Annotations were stored in JSON format.  
Each instrument instance contains the required number of keypoints, with missing ones marked as ``not available''.

\subsection{Generation of challenge dataset}
\label{methods:challenge}

The dataset served as the training resource of the PhaKIR Challenge at MICCAI 2024.  
For each video, the raw video file, cut indices, and annotations for all tasks were bundled into a single archive.  
To reduce storage requirements, complete videos were distributed together with a frame extraction script, allowing users to control frame compression.  

\section*{VALIDATION AND QUALITY}

In the following, we describe the measures taken to ensure the high quality of the dataset, followed by a discussion of its potential limitations.

\subsection{Validation of annotations}

The quality of the dataset was ensured through multiple validation steps, described in detail below.
We first outline the general validation strategy, then describe task-specific procedures, and finally summarize additional validation measures.

\subsubsection{General approach}
\label{validation:general}

At the beginning of the annotation process, the medical annotation team was provided with a detailed annotation protocol containing task-specific guidelines and illustrative examples of various scenarios.
This protocol was developed jointly with medical experts and continuously refined during the annotation process.

A single annotator first annotated each video.
After completion, the annotations underwent a renewed inspection by the same annotator to correct potential oversights.
Subsequently, two additional team members verified the annotations sequentially, following the protocol.
Thus, each video was reviewed three times after its initial annotation.

Given that the three presented tasks are less complex to annotate compared to other biomedical tasks, such as tumor segmentation or tissue type classification, and considering the large amount of annotated data, we adopted this sequential validation strategy instead of parallel multi-annotation with label merging mechanisms.
This ensured both high annotation quality and efficiency.

\subsubsection{Surgical phase recognition}
\label{validation:phase}

\definecolor{argonbeamer}{RGB}{60, 50, 50}
\definecolor{bipolar}{RGB}{89, 134, 179}
\definecolor{bluntgrasper}{RGB}{128, 128, 128}
\definecolor{bluntgraspercurved}{RGB}{200, 102, 235}
\definecolor{bluntgrasperspec}{RGB}{179, 102, 235}

\definecolor{clipapplicator}{RGB}{0, 0, 255}
\definecolor{dissectionhook}{RGB}{80, 140, 0}
\definecolor{drainage}{RGB}{255, 100, 0}
\definecolor{grasper}{RGB}{255, 130, 0}
\definecolor{hfcoag}{RGB}{255, 0, 153}

\definecolor{hookclamp}{RGB}{0, 80, 80}
\definecolor{needle}{RGB}{204, 153, 153}
\definecolor{overholt}{RGB}{255, 200, 170}
\definecolor{palpation}{RGB}{255, 102, 255}
\definecolor{pe}{RGB}{30, 144, 1}

\definecolor{scissor}{RGB}{255, 255, 0}
\definecolor{sponge}{RGB}{40, 120, 80}
\definecolor{suction}{RGB}{153, 0, 204}
\definecolor{trocar}{RGB}{153, 102, 0}

\begin{table*}[t]
    \small
    \caption{Designation of the surgical instrument types (Inst.) together with the color codes used in the segmentation masks (RGB) and visualization of the respective color (Vis.), sorted in ascending alphabetical order according to the instrument names. The displayed R and G values are derived from the instrument classes. For the B channel, the values of the instances that appear for the first time in a video for an instrument are presented.}
    \centering
    \begin{tabularx}{1.0\textwidth}{
l 
>{\centering\arraybackslash}m{2.55cm} 
>{\centering\arraybackslash}m{2.55cm} 
>{\centering\arraybackslash}m{2.55cm} 
>{\centering\arraybackslash}m{2.55cm} 
>{\centering\arraybackslash}m{2.55cm}
}
\toprule
\textbf{Inst.} & Argonbeamer & Bipolar-Clamp & Blunt-Grasper & Blunt-Grasper-Curved & Blunt-Grasper-Spec \\
\textbf{RGB} & \texttt{[060,050,050]} & \texttt{[089,134,179]} & \texttt{[128,128,128]} & \texttt{[200,102,235]} & \texttt{[179,102,235]} \\
\textbf{Vis.} & 
\tikz \fill[argonbeamer] (0,0) rectangle (2.25cm,0.4cm); &
\tikz \fill[bipolar] (0,0) rectangle (2.25cm,0.4cm); &
\tikz \fill[bluntgrasper] (0,0) rectangle (2.25cm,0.4cm); &
\tikz \fill[bluntgraspercurved] (0,0) rectangle (2.25cm,0.4cm); &
\tikz \fill[bluntgrasperspec] (0,0) rectangle (2.25cm,0.4cm); \\
\midrule

\textbf{Inst.} & Clip-Applicator & Dissection-Hook & Drainage & Grasper & HF-Coag.-Probe \\
\textbf{RGB} & \texttt{[000,000,255]} & \texttt{[080,140,000]} & \texttt{[255,100,000]} & \texttt{[255,130,000]} & \texttt{[255,000,153]} \\
\textbf{Vis.} &
\tikz \fill[clipapplicator] (0,0) rectangle (2.25cm,0.4cm); &
\tikz \fill[dissectionhook] (0,0) rectangle (2.25cm,0.4cm); &
\tikz \fill[drainage] (0,0) rectangle (2.25cm,0.4cm); &
\tikz \fill[grasper] (0,0) rectangle (2.25cm,0.4cm); &
\tikz \fill[hfcoag] (0,0) rectangle (2.25cm,0.4cm); \\
\midrule

\textbf{Inst.} & Hook-Clamp & Needle-Probe & Overholt & Palpation-Probe & PE-Forceps \\
\textbf{RGB} & \texttt{[000,080,080]} & \texttt{[204,153,153]} & \texttt{[255,200,170]} & \texttt{[255,102,255]} & \texttt{[030,144,001]} \\
\textbf{Vis.} &
\tikz \fill[hookclamp] (0,0) rectangle (2.25cm,0.4cm); &
\tikz \fill[needle] (0,0) rectangle (2.25cm,0.4cm); &
\tikz \fill[overholt] (0,0) rectangle (2.25cm,0.4cm); &
\tikz \fill[palpation] (0,0) rectangle (2.25cm,0.4cm); &
\tikz \fill[pe] (0,0) rectangle (2.25cm,0.4cm); \\
\midrule

\textbf{Inst.} & Scissor & Sponge-Clamp & Suction-Rod & Trocar-Tip & --- \\
\textbf{RGB} & \texttt{[255,255,000]} & \texttt{[040,120,080]} & \texttt{[153,000,204]} & \texttt{[153,102,000]} & --- \\
\textbf{Vis.} &
\tikz \fill[scissor] (0,0) rectangle (2.25cm,0.4cm); &
\tikz \fill[sponge] (0,0) rectangle (2.25cm,0.4cm); &
\tikz \fill[suction] (0,0) rectangle (2.25cm,0.4cm); &
\tikz \fill[trocar] (0,0) rectangle (2.25cm,0.4cm); &
--- \\

\bottomrule
\end{tabularx}

\label{tab:segmentation_colors}
\end{table*}

For surgical phase recognition, the annotated timestamps were repeatedly checked according to the procedure described above.
In addition, plausibility checks were performed by verifying the presence of instruments in corresponding phases, as some instrument classes are only expected to occur in specific phases.
Unlike the instrument instance segmentation task, no quantitative correction statistics were recorded; validation was based on repeated review and cross-checking against the surgical workflow.

\subsubsection{Instrument instance segmentation}
\label{validation:segmentation}

For this task, the verification of the original annotations was supplemented by a visual inspection.
The annotations were visualized (see the first row in Fig.~\ref{fig:annotations}) and combined side by side with the original frames to create a new composite image, allowing for a direct comparison of the input frames and segmentation labels.
A correction video was generated from these combined images and subjected to the three-stage validation procedure.
By monitoring the color-coded instrument representations over time, inconsistencies could be detected, such as changes in the assigned class of an instrument between consecutive frames.

Between the second and third review passes, correction rates were quantified. 
At that point, the multi-instance multi-class dice score (DSC) was applied analogously to the PhaKIR challenge evaluation (see~\cite{rueckert2025comparative}), which revealed an annotation agreement of 83.64\%.
The main source of error was the classification of surgical instruments, which then entered the result with a DSC of zero per instrument instance. 
This observation suggests that many discrepancies had already been eliminated in the earlier review phases, so that only minor improvements were necessary for the final validation round.

\subsubsection{Instrument keypoint estimation}
\label{validation:keypoint}

Validation of keypoint annotations was likewise based on visual inspection.
Annotations were performed in CVAT on top of the segmentation labels, thereby implicitly validating the segmentations and classifications.
Keypoints were annotated by a team member who did not contribute to the original segmentations or correction runs, ensuring an independent annotation and review.
Similar to the segmentation validation, input images were combined with the keypoint annotations (see the fourth row in Fig.~\ref{fig:annotations}), compiled into videos, and subjected to the three-stage validation procedure.
For this task, no quantitative correction rates were tracked; quality assurance relied on repeated visual inspection and temporal consistency checks under occlusion.

\subsubsection{Community validation}
\label{validation:further}

\begin{figure*}[t!]
\centering
\small

\definecolor{futureblue}{RGB}{100, 100, 255}
\definecolor{futuredim}{RGB}{150, 150, 150}

\begin{tikzpicture}[
    folder/.style={rectangle, draw=none, minimum width=0.0cm, minimum height=0.7cm, text centered},
    file/.style={rectangle, draw=none, minimum width=1.8cm, minimum height=0.01cm, text centered},
    fileBlue/.style={rectangle, draw=none, fill=none, text=futureblue, minimum width=1.8cm, minimum height=0.01cm, text centered},
    arrow/.style={-{Latex}, thick},
    level 1/.style={sibling distance=5cm},
    level 2/.style={sibling distance=3cm},
    node distance=0.1cm
]

\node[folder] (vid01) {\begin{tabular}{c} \includegraphics[width=0.03\textwidth]{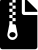} \\ Video\_01\\.zip \end{tabular}};

\node[folder, below=0.2cm of vid01] (vid02) {\begin{tabular}{c} \includegraphics[width=0.03\textwidth]{icon_file_zip.pdf} \\ Video\_02\\.zip \end{tabular}};

\node[folder, right=-0.5cm of vid02] (vid03) {\begin{tabular}{c} \includegraphics[width=0.03\textwidth]{icon_file_zip.pdf} \\ Video\_03\\.zip \end{tabular}};

\node[folder, below=0.2cm of vid02] (vid04) {\begin{tabular}{c} \includegraphics[width=0.03\textwidth]{icon_file_zip.pdf} \\ Video\_04\\.zip \end{tabular}};

\node[folder, right=-0.5cm of vid04] (vid05) {\begin{tabular}{c} \includegraphics[width=0.03\textwidth]{icon_file_zip.pdf} \\ Video\_05\\.zip \end{tabular}};

\node[folder, below=0.2cm of vid04] (vid07) {\begin{tabular}{c} \includegraphics[width=0.03\textwidth]{icon_file_zip.pdf} \\ Video\_07\\.zip \end{tabular}};

\node[folder, right=-0.5cm of vid07] (vid11) {\begin{tabular}{c} \includegraphics[width=0.03\textwidth]{icon_file_zip.pdf} \\ Video\_11\\.zip \end{tabular}};

\node[folder, below=0.2cm of vid07] (vid13) {\begin{tabular}{c} \includegraphics[width=0.03\textwidth]{icon_file_zip.pdf} \\ Video\_13\\.zip \end{tabular}};

\node[folder, right=-0.5cm of vid13] (split) {\begin{tabular}{c} \includegraphics[width=0.03\textwidth]{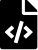} \\ split\_video\_\\in\_frames.py \end{tabular}};

\node[folder, right=2.0cm of vid01, yshift=-0.0cm] (v01_01) {\begin{tabular}{c} \includegraphics[width=0.03\textwidth]{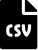} \\ Video\_01\_\\Cuts.csv \end{tabular}};

\node[folder, below=0.2cm of v01_01] (v01_02) {\begin{tabular}{c} \includegraphics[width=0.03\textwidth]{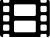} \\ Video\_01\hphantom{--}\\.mp4 \end{tabular}};

\node[folder, below=0.2cm of v01_02] (v01_03) {\begin{tabular}{c} \includegraphics[width=0.03\textwidth]{icon_file_code.pdf} \\ Video\_01\_\\Keypoints.json \end{tabular}};

\node[folder, below=0.2cm of v01_03] (v01_04) {\begin{tabular}{c} \includegraphics[width=0.03\textwidth]{icon_file_csv.pdf} \\ Video\_01\_\\Phases.csv \end{tabular}};

\node[folder, below=0.2cm of v01_04] (v01_05) {\begin{tabular}{c} \includegraphics[width=0.03\textwidth]{icon_file_zip.pdf} \\ Video\_01\_\\Masks.zip \end{tabular}};

  \path let 
    \p1 = (v01_05.south),
    \p2 = (v01_01.north)
  in
    coordinate (rectSW) at ($(\x1,\y1) + (-1.2cm,0.0cm)$)
    coordinate (rectNE) at ($(\x2,\y2) + (1.2cm,0.0cm)$);

   \draw[black, thick, fill=none, rounded corners]
     (rectSW) rectangle (rectNE);

\node[fileBlue, right=0.8cm of v01_02, yshift=2.0cm] (v01_02_01) {
\begin{tabular}{@{}m{0.025\textwidth}@{} m{0.05\textwidth}@{}}
  \includegraphics[width=0.03\textwidth]{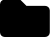} 
  &
  \centering 0
\end{tabular}
};

\node[fileBlue, below=0.1cm of v01_02_01] (v01_02_02) {
\begin{tabular}{@{}m{0.03\textwidth}@{} m{0.05\textwidth}@{}}
  \includegraphics[width=0.03\textwidth]{icon_folder_closed.pdf} 
  &
  \centering 1000
\end{tabular}
};

\node[fileBlue, below=0.1cm of v01_02_02] (v01_02_03) {
\begin{tabular}{@{}m{0.03\textwidth}@{} m{0.05\textwidth}@{}}
  \includegraphics[width=0.03\textwidth]{icon_folder_closed.pdf} 
  &
  \centering \dots
\end{tabular}
};

\node[fileBlue, below=0.1cm of v01_02_03] (v01_02_04) {
\begin{tabular}{@{}m{0.035\textwidth}@{} m{0.05\textwidth}@{}}
  \includegraphics[width=0.03\textwidth]{icon_folder_closed.pdf} 
  &
  \centering 58000
\end{tabular}
};

\node[file, below=3.0cm of v01_02_04] (v01_05_01) {
\begin{tabular}{@{}m{0.025\textwidth}@{} m{0.05\textwidth}@{}}
  \includegraphics[width=0.03\textwidth]{icon_folder_closed.pdf} 
  &
  \centering 0
\end{tabular}
};

\node[file, below=0.1cm of v01_05_01] (v01_05_02) {
\begin{tabular}{@{}m{0.03\textwidth}@{} m{0.05\textwidth}@{}}
  \includegraphics[width=0.03\textwidth]{icon_folder_closed.pdf} 
  &
  \centering 1000
\end{tabular}
};

\node[file, below=0.1cm of v01_05_02] (v01_05_03) {
\begin{tabular}{@{}m{0.03\textwidth}@{} m{0.05\textwidth}@{}}
  \includegraphics[width=0.03\textwidth]{icon_folder_closed.pdf} 
  &
  \centering \dots
\end{tabular}
};

\node[file, below=0.1cm of v01_05_03] (v01_05_04) {
\begin{tabular}{@{}m{0.035\textwidth}@{} m{0.05\textwidth}@{}}
  \includegraphics[width=0.03\textwidth]{icon_folder_closed.pdf} 
  &
  \centering 58000
\end{tabular}
};

\node[fileBlue, right=1.3cm of v01_02_03, yshift=3.0cm] (v01_02_03_01) {
\begin{tabular}{c} \dots \end{tabular}
};

\node[fileBlue, below=0.1cm of v01_02_03_01] (v01_02_03_02) {
\begin{tabular}{c} \includegraphics[width=0.1\textwidth]{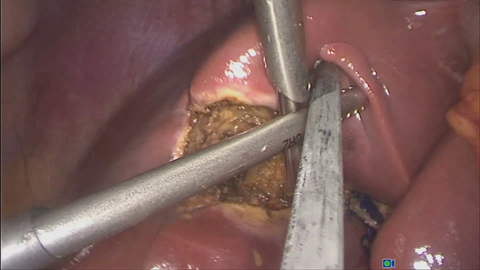} \\ frame\_\\055649.png \end{tabular}
};

\node[fileBlue, below=0.1cm of v01_02_03_02] (v01_02_03_03) {
\begin{tabular}{c} \includegraphics[width=0.1\textwidth]{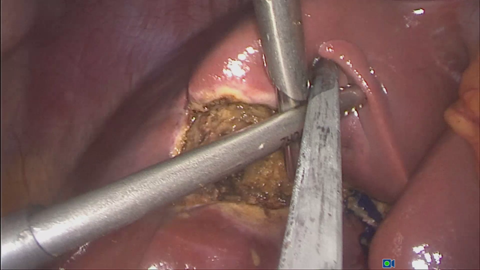} \\ frame\_\\055650.png \end{tabular}
};

\node[fileBlue, below=0.1cm of v01_02_03_03] (v01_02_03_04) {
\begin{tabular}{c} \dots \end{tabular}
};

\draw[decorate,decoration={brace,amplitude=15pt}, draw=futureblue, line width=1pt, xshift=-30pt]
  (v01_02_03_04.south west) -- (v01_02_03_01.north west)
  node[text=futureblue, midway,xshift=-23pt,rotate=90] (frame_brace) { sequential frames};

\node[file, below=0.5cm of v01_02_03_04] (v01_05_03_01) {
\begin{tabular}{c} \dots \end{tabular}
};

\node[file, below=0.1cm of v01_05_03_01] (v01_05_03_02) {
\begin{tabular}{c} \includegraphics[width=0.1\textwidth]{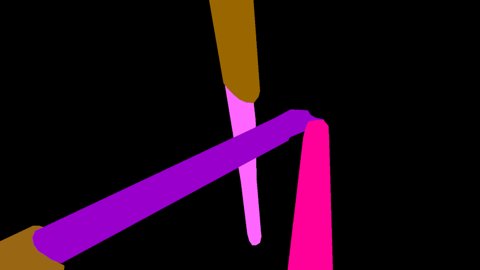} \\ 055625\_\\mask.png \end{tabular}
};

\node[file, below=0.1cm of v01_05_03_02] (v01_05_03_03) {
\begin{tabular}{c} \includegraphics[width=0.1\textwidth]{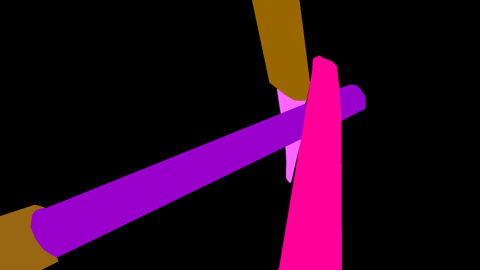} \\ 055650\_\\mask.png \end{tabular}
};

\node[file, below=0.1cm of v01_05_03_03] (v01_05_03_04) {
\begin{tabular}{c} \dots \end{tabular}
};

\draw[decorate,decoration={brace,amplitude=15pt}, line width=1pt, xshift=-30pt]
  (v01_05_03_04.south west) -- (v01_05_03_01.north west)
  node[midway,xshift=-23pt,rotate=90] (mask_brace) { interval masks};

\draw[arrow] (vid01) -- ($(v01_01)-(1.5,0)$);

\draw[arrow, draw=futureblue, shorten <=5mm] (v01_02) -- (v01_02_01.west);
\draw[arrow, draw=futureblue] (v01_02) -- (v01_02_02.west);
\draw[arrow, draw=futureblue] (v01_02) -- (v01_02_03.west);
\draw[arrow, draw=futureblue] (v01_02) -- (v01_02_04.west);

\draw[arrow, shorten <= 1mm] (v01_05) -- (v01_05_01.west);
\draw[arrow] (v01_05) -- (v01_05_02.west);
\draw[arrow] (v01_05) -- (v01_05_03.west);
\draw[arrow] (v01_05) -- (v01_05_04.west);

\draw[arrow, draw=futureblue] ([xshift=-3mm]v01_02_03.east) -- ++(0.50,0);

\draw[arrow] ([xshift=-3mm]v01_05_03.east) -- ++(0.50,0);

\end{tikzpicture}

\caption{Structure of the dataset. The videos available are displayed on the left, the middle column shows the individual elements of each video, and the fine-grained folder structure for the frames and masks is displayed on the right. The elements marked in blue are generated after the frame extraction using the provided script.}
\label{fig:dataset_structure}
\end{figure*}

In addition to the measures described above, the dataset was employed as the training part in the PhaKIR challenge, providing implicit quality assurance through large-scale community use.
As reported in~\cite{rueckert2025comparative}, it was downloaded and inspected by 66 registered teams worldwide, representing a much larger number of individual participants.
Rare annotation errors identified and reported by participants during the challenge were corrected, and the dataset was continuously updated.
Importantly, no systematic errors were reported during the challenge, further confirming the robustness of the final dataset.

\subsection{Limitations of the dataset}

Several limitations should be considered when using the presented dataset.
A key strength of the dataset is the inclusion of complete intervention sequences from three different medical centers, which allows for temporal modeling and inter-institutional comparison.
Nevertheless, the number of procedures is limited and may not capture the full range of variability of surgical practice.
All recordings depict laparoscopic cholecystectomies, so other types of procedures are not represented.
Moreover, the data originate exclusively from German medical centers, and surgical practices for cholecystectomies may differ in other countries.
Finally, the dataset contains only RGB video; no other multimodal information is included.
Users should take these factors into account when interpreting results or developing models based on the dataset.

\section*{RECORDS AND STORAGE} 
 
The organization of the PhaKIR dataset is illustrated in Figure~\ref{fig:dataset_structure}, which shows nine files in the root level.  
Eight training videos are provided as compressed zip archives, together with the Python script for frame extraction.  
The video archives are not numbered sequentially, as intermediate sequences were reserved for the PhaKIR test dataset.  

Each video archive follows the same internal structure and contains five files, where the placeholder \texttt{xx} denotes the video identifier.
At this level, videos, annotations, and auxiliary metadata are kept together to ensure direct correspondence between raw data and labels.
The design follows FAIR data principles~\cite{wilkinson2016Fair} by relying on open formats (CSV, JSON, PNG, MP4), consistent file naming, and a reproducible directory structure.
\begin{itemize}
    \item \texttt{Video\_xx\_Cuts.csv}: Frame indices marking segments that were removed for anonymization.
    \item \texttt{Video\_xx.mp4}: The surgical video at 25 fps and resolution $1920 \times 1080$.  
    Frames can be extracted using the provided script \texttt{split\_video\_in\_frames.py}.
    Extracted frames are stored in subfolders of 1000 images, with filenames zero-padded to six digits (see blue-colored text in Figure~\ref{fig:dataset_structure}.
    This structure ensures efficient storage and fast access.
    \item \texttt{Video\_xx\_Keypoints.json}: Keypoint annotations for all surgical instruments, including visibility states, stored in JSON format.  
    \item \texttt{Video\_xx\_Phases.csv}: Frame-level annotations of surgical phases throughout the intervention.  
    \item \texttt{Video\_xx\_Masks.zip}: Archive of segmentation masks encoded in color channels.  
    Masks are organized into subfolders of 1000 files each, mirroring the extracted frame structure and filename convention.  
    The color coding of instrument classes and instances is summarized in Table~\ref{tab:segmentation_colors}.  
\end{itemize}
For clarity, Table~\ref{tab:data_records} summarizes the contents of each file, their format, and specific conventions.
This compact overview is intended as a quick reference for users when navigating the dataset.

\balance
\section*{INSIGHTS AND NOTES} 

The proposed dataset builds upon and extends two previously published datasets, namely the PhaKIR Challenge dataset~\cite{rueckert2025comparative} and the HeiChole Challenge dataset~\cite{wagner2023comparative}, both of which were released under the CC-BY-NC-SA license~\cite{CC-BY-NC-SA-4-0}. 
Accordingly, the present dataset is distributed under the same licensing terms and is publicly accessible upon request at \url{https://zenodo.org/records/15740619}. 
When using this dataset, either in full or in part, users are required to cite this dataset publication, the corresponding challenge paper~\cite{rueckert2025comparative}, and the HeiChole challenge publication~\cite{wagner2023comparative} in any resulting scientific work.

Beyond licensing considerations, the dataset is intended to support a broad range of research in surgical data science.
Its unified annotations enable both single-task and multi-task approaches, allowing the benchmarking of surgical phase recognition, instrument segmentation, and instrument keypoint estimation within a single dataset.
Researchers may replicate the setup of the PhaKIR Challenge at MICCAI 2024, where all eight videos were released for training and an independent set was reserved for testing.
Since the challenge test set is not publicly available, we recommend that users create validation splits at the video level rather than at the frame level.
Possible strategies include leave-one-video-out or leave-one-hospital-out protocols, which allow for the evaluation of temporal modeling and generalization across institutions.

Video data are provided in MP4 format together with a frame extraction script.
Users may choose to work at the native 25 fps for phase recognition, or at reduced frame rates (e.g., 1 fps) for segmentation and keypoint tasks, depending on computational resources.
The directory structure with subfolders of 1000 frames and zero-padded filenames facilitates efficient storage management and straightforward alignment of videos with annotations.

The dataset can also be combined with other publicly available resources.
For example, Cholec80~\cite{twinanda2016endonet} provides additional phase annotations, while EndoVis challenge datasets~\cite{allan2017robotic}, \cite{allan2018robotic} include complementary segmentation tasks.
Such combinations enable cross-dataset evaluation, transfer learning, and studies on domain shift across institutions and recording setups.

\section*{SOURCE CODE AND SCRIPTS} 

\begin{table*}[t!]
    \small
    \centering
    \caption{Summary of files contained in each video archive, their content, format, and conventions.}
    \label{tab:data_records}
    \renewcommand{\arraystretch}{1.0}
    \begin{tabularx}{\textwidth}{p{0.22\textwidth} p{0.175\textwidth} p{0.07\textwidth} p{0.4\textwidth}}
        \toprule
        \textbf{Filename} & \textbf{Content} & \textbf{Format} & \textbf{Notes} \\
        \midrule
        \texttt{Video\_xx\_Cuts.csv} & Anonymization cut indices & CSV & Aligns with video frame numbering. \\
        \texttt{Video\_xx.mp4} & Surgical video & MP4 & 25 fps, $1920 \times 1080$; 1000-frame subfolders; zero-padded filenames. \\
        \texttt{Video\_xx\_Keypoints.json} & Keypoint annotations & JSON & Per frame $\rightarrow$ per instrument $\rightarrow$ coordinates $+$ visibility. \\
        \texttt{Video\_xx\_Phases.csv} & Surgical phase labels & CSV & Frame index $+$ phase label (7 phases $+$ \textit{undefined}). \\
        \texttt{Video\_xx\_Masks.zip} & Segmentation masks & ZIP (PNG) & Subfolders of 1000; filenames match frames; color encoding in Table~\ref{tab:segmentation_colors}. \\
        \bottomrule
    \end{tabularx}
\end{table*}

The script \texttt{split\_video\_in\_frames.py} is provided at the top level of the dataset archive on Zenodo and can be used to split the video sequences into individual frames.
It also allows control over the compression rate of the resulting frames, depending on the available storage capacity.
The script is written in Python and does not require external dependencies beyond standard libraries.
It is released under the same license as the dataset (CC-BY-NC-SA) to ensure reproducibility and open accessibility.
No further scripts are required for using the dataset, which keeps the workflow minimal and transparent.

\section*{ACKNOWLEDGEMENTS}
We would like to express our sincere thanks for supporting the annotation process to Michel Xiao, Lukas Beckendorf, Egzona Humolli, and Regine Hartwig, who collaborated as student assistants with the MITI research group during this period.
This work was supported by the Bavarian Research Foundation [BFS, grant number AZ-1506-21], and the Bavarian Academic Forum [BayWISS]. 
Funding for open access publication was provided by OTH Regensburg.
\\
\text{\hspace{1em}} T.~R. conceptualized the dataset, wrote the labeling protocol, supported the annotation process, supported the technical realization of the annotation process, and wrote the manuscript.
R.~M. and D.~R. supported the conceptualization of the dataset and the annotation process.
L.~K. supported the technical realization of the annotation process.
M.~G. supported the conceptualization of the dataset.
D.~R. supervised the project.
H.~F. conceptualized the dataset, acquired the data, supervised and coordinated the annotation process, created the annotations, verified the annotations, supported the writing of the labeling protocol, and supervised the project.
D.~W. conceptualized the dataset, acquired the data, and supervised the project.
C.~P. conceptualized the dataset, supervised the project, and wrote the manuscript.
All authors read and approved the final version of the manuscript.
\\
\text{\hspace{1em}} The article authors have declared no conflicts of interest.
\\
\bibliographystyle{IEEEtran}

\bibliography{main}

\begin{thebibliography}{10}
\providecommand{\url}[1]{#1}
\csname url@samestyle\endcsname
\providecommand{\newblock}{\relax}
\providecommand{\bibinfo}[2]{#2}
\providecommand{\BIBentrySTDinterwordspacing}{\spaceskip=0pt\relax}
\providecommand{\BIBentryALTinterwordstretchfactor}{4}
\providecommand{\BIBentryALTinterwordspacing}{\spaceskip=\fontdimen2\font plus
\BIBentryALTinterwordstretchfactor\fontdimen3\font minus
  \fontdimen4\font\relax}
\providecommand{\BIBforeignlanguage}[2]{{%
\expandafter\ifx\csname l@#1\endcsname\relax
\typeout{** WARNING: IEEEtran.bst: No hyphenation pattern has been}%
\typeout{** loaded for the language `#1'. Using the pattern for}%
\typeout{** the default language instead.}%
\else
\language=\csname l@#1\endcsname
\fi
#2}}
\providecommand{\BIBdecl}{\relax}
\BIBdecl

\bibitem{darzi2002recent}
A.~Darzi and S.~Mackay, ``Recent advances in minimal access surgery,''
  \emph{Bmj}, vol. 324, no. 7328, pp. 31--34, 2002.

\bibitem{de2018minimally}
T.~de~Rooij \emph{et~al.}, ``Minimally invasive vs. open distal pancreatectomy
  (leopard): Multicenter patient-blinded randomized controlled trial,''
  \emph{HPB}, vol.~20, pp. S293--S294, 2018.

\bibitem{van2012robot}
P.~C. van~der Sluis \emph{et~al.}, ``Robot-assisted minimally invasive
  thoraco-laparoscopic esophagectomy versus open transthoracic esophagectomy
  for resectable esophageal cancer, a randomized controlled trial (robot
  trial),'' \emph{Trials}, vol.~13, no.~1, p. 230, 2012.

\bibitem{haidegger2022robot}
T.~Haidegger, S.~Speidel, D.~Stoyanov, and R.~M. Satava, ``Robot-assisted
  minimally invasive surgery—surgical robotics in the data age,''
  \emph{Proceedings of the IEEE}, vol. 110, no.~7, pp. 835--846, 2022.

\bibitem{maier2022surgical}
L.~Maier-Hein \emph{et~al.}, ``Surgical data science--from concepts toward
  clinical translation,'' \emph{Medical image analysis}, vol.~76, p. 102306,
  2022.

\bibitem{rueckert2024methods}
T.~Rueckert, D.~Rueckert, and C.~Palm, ``Methods and datasets for segmentation
  of minimally invasive surgical instruments in endoscopic images and videos:
  {A} review of the state of the art,'' \emph{Comput. Biol. Med.}, vol. 169, p.
  107929, 2024.

\bibitem{twinanda2016endonet}
A.~P. Twinanda, S.~Shehata, D.~Mutter, J.~Marescaux, M.~De~Mathelin, and
  N.~Padoy, ``Endonet: a deep architecture for recognition tasks on
  laparoscopic videos,'' \emph{IEEE transactions on medical imaging}, vol.~36,
  no.~1, pp. 86--97, 2016.

\bibitem{stauder2016tum}
R.~Stauder, D.~Ostler, M.~Kranzfelder, S.~Koller, H.~Feu{\ss}ner, and N.~Navab,
  ``The tum lapchole dataset for the m2cai 2016 workflow challenge,''
  \emph{arXiv preprint arXiv:1610.09278}, 2016.

\bibitem{wagner2023comparative}
M.~Wagner \emph{et~al.}, ``Comparative validation of machine learning
  algorithms for surgical workflow and skill analysis with the heichole
  benchmark,'' \emph{Medical image analysis}, vol.~86, p. 102770, 2023.

\bibitem{bodenstedt2015comparative}
S.~Bodenstedt \emph{et~al.}, ``Comparative evaluation of instrument
  segmentation and tracking methods in minimally invasive surgery,''
  \emph{ArXiv preprint}, vol. abs/1805.02475, 2018.

\bibitem{allan2017robotic}
M.~Allan \emph{et~al.}, ``2017 robotic instrument segmentation challenge,''
  \emph{ArXiv preprint}, vol. abs/1902.06426, 2019.

\bibitem{allan2018robotic}
------, ``2018 robotic scene segmentation challenge,'' \emph{ArXiv preprint},
  vol. abs/2001.11190, 2020.

\bibitem{ross2019comparative}
T.~Ro{\ss} \emph{et~al.}, ``Comparative validation of multi-instance instrument
  segmentation in endoscopy: Results of the {ROBUST-MIS} 2019 challenge,''
  \emph{Medical Image Anal.}, vol.~70, p. 101920, 2021.

\bibitem{zia2021objective}
A.~Zia \emph{et~al.}, ``Objective surgical skills assessment and tool
  localization: Results from the {MICCAI} 2021 simsurgskill challenge,''
  \emph{ArXiv preprint}, vol. abs/2212.04448, 2022.

\bibitem{psychogyios2022sar}
D.~Psychogyios \emph{et~al.}, ``{SAR-RARP50:} segmentation of surgical
  instrumentation and action recognition on robot-assisted radical
  prostatectomy challenge,'' \emph{ArXiv preprint}, vol. abs/2401.00496, 2024.

\bibitem{nwoye2022cholec}
C.~I. Nwoye \emph{et~al.}, ``Cholectriplet2022: Show me a tool and tell me the
  triplet - an endoscopic vision challenge for surgical action triplet
  detection,'' \emph{Medical Image Anal.}, vol.~89, p. 102888, 2023.

\bibitem{zia2022surgical}
A.~Zia \emph{et~al.}, ``Surgical tool classification and localization: results
  and methods from the {MICCAI} 2022 surgtoolloc challenge,'' \emph{ArXiv
  preprint}, vol. abs/2305.07152, 2023.

\bibitem{malpani2023synthetic}
A.~Malpani and K.~M. Glock, ``{Syn-ISS:} synthetic data for instrument
  segmentation in surgery,'' 2023,
  \url{https://www.synapse.org/Synapse:syn50908388/wiki/620516} [Accessed:
  2025-10-15].

\bibitem{bodenstedt2023}
S.~Bodenstedt, A.~Jenke, S.~Speidel, M.~D. Martin~Wagner, and A.~Tabibian,
  ``Sims: Surgical instrument multi-domain segmentation challenge,'' 2023,
  \url{https://www.synapse.org/Synapse:syn47193563/wiki/620035} [Accessed:
  2025-10-15].

\bibitem{rueckert2025comparative}
T.~Rueckert \emph{et~al.}, ``Comparative validation of surgical phase
  recognition, instrument keypoint estimation, and instrument instance
  segmentation in endoscopy: Results of the {PhaKIR} 2024 challenge,''
  \emph{arXiv preprint arXiv:2507.16559}, 2025.

\bibitem{rueckert2025dataset}
\BIBentryALTinterwordspacing
------, ``{PhaKIR Dataset - Surgical Procedure Phase, Keypoint, and Instrument
  Recognition},'' 2025. [Online]. Available:
  \url{https://doi.org/10.5281/zenodo.15740619}
\BIBentrySTDinterwordspacing

\bibitem{MRI_TUM}
``{TUM Klinikum Rechts der Isar – Universitätsklinikum der Technischen
  Universität München},'' \url{https://www.mri.tum.de/}, accessed: 15 October
  2025.

\bibitem{UKHD}
``{Universit\"atsklinikum Heidelberg},''
  \url{https://www.klinikum.uni-heidelberg.de/}, accessed: 15 October 2025.

\bibitem{MeinKrankenhaus2030}
``{meinKrankenhaus2030 – Krankenhaus Weilheim-Schongau},''
  \url{https://www.meinkrankenhaus2030.de/}, accessed: 15 October 2025.

\bibitem{cvatai2023cvat}
\BIBentryALTinterwordspacing
C.~Corporation, ``Computer vision annotation tool (cvat),'' 2023, version
  2.25.0, MIT License. [Online]. Available: \url{https://cvat.ai/}
\BIBentrySTDinterwordspacing

\bibitem{coco_keypoints}
``Coco dataset keypoints evaluation,''
  \url{https://cocodataset.org/\#keypoints-eval}, 2025, accessed: 15 October
  2025.

\bibitem{wilkinson2016Fair}
M.~Wilkinson \emph{et~al.}, ``The fair guiding principles for scientific data
  management and stewardship,'' \emph{Scientific Data}, vol.~3, 03 2016.

\bibitem{CC-BY-NC-SA-4-0}
{Creative Commons}, ``{Creative Commons Attribution-NonCommercial-ShareAlike
  4.0 International Public License},''
  \url{https://creativecommons.org/licenses/by-nc-sa/4.0/}, 2013, accessed:
  2025-10-15.

\end{thebibliography}

\end{document}